\documentclass[11pt]{article}
\usepackage{naacl2001}

\title{Machine Learning with Lexical Features: \\
The Duluth Approach to \textsc{Senseval-2}}

\author{Ted Pedersen\\
Department of Computer Science\\
University of Minnesota, Duluth
\\Duluth, MN 55812 USA\\
\tt{tpederse@d.umn.edu}} 

\begin{document}
\maketitle      

\begin{abstract}
This paper describes the sixteen Duluth entries in the \textsc{Senseval-2}
comparative exercise among word sense disambiguation systems. There were
eight pairs of Duluth systems entered in the Spanish and English lexical 
sample tasks. These are all based on standard machine learning algorithms
that induce classifiers from sense-tagged training text where the context 
in which ambiguous words occur are represented by simple lexical features. 
These are highly portable, robust methods that can serve as a foundation
for more tailored approaches. 
\end{abstract}

\section{Introduction}

The Duluth systems in \textsc{Senseval-2} take a supervised learning approach
to the Spanish and English lexical sample tasks. They
learn decision trees and Naive Bayesian classifiers from sense-tagged 
training examples where the context in which an ambiguous word 
occurs is represented by lexical features. These include unigrams
and bigrams that occur anywhere in the context, and co-occurrences within 
just a few words of the target word. These are the only types of
features used. There are no syntactic features, nor is the structure or 
content of WordNet employed. As a result these systems are highly 
portable, and can serve as a foundation for systems that are 
tailored to particular languages and sense inventories.

The word sense disambiguation literature provides ample evidence that 
many different kinds of features contribute to the resolution of
word meaning. These include part--of--speech, morphology, verb--object 
relationships, selectional restrictions, lexical features, etc. When 
used in combination it is often unclear to what degree each type of 
feature contributes to overall performance. It is also unclear to what
extent adding new features allows for the disambiguation of previously 
unresolvable test instances. One of the long term objectives of our 
research is to determine which  types of features are complementary and 
cover increasing numbers of test instances as they are added to a 
representation of context. 

\section{Experimental Methodology}

The training and test data for the English and Spanish lexical 
sample tasks is split into separate training and test files per word. A 
supervised learning algorithm induces a classifier from the training
examples for a word, which is then used to assign sense tags to the test 
instances for that word. 

The context in which an ambiguous word occurs is represented 
by lexical features that are identified using the Bigram Statistics
Package (BSP) version 0.4. This is free software that extracts unigrams 
and bigrams from text using a variety of statistical methods. Each unigram 
or bigram that is identified in the training data is treated as a binary 
feature that indicates whether or not it occurs in the context of the word 
being disambiguated. The free software package SenseTools (version 
0.1) converts training  and test data into a feature vector
representation, based on the output from BSP. This becomes the input to 
the Weka suite of supervised learning algorithms. Weka induces classifiers 
from the training  examples and applies the sense tags to the test 
instances. 

The same software is used for the English and Spanish text. 
BSP and SenseTools are written in Perl and are freely available 
from www.d.umn.edu/\~{}tpederse/code.html.  Weka is written in Java and
is freely available from  www.cs.waikato.ac.nz/\~{}ml. 

\section{System Descriptions}

There were eight pairs of Duluth systems in the English and Spanish
lexical sample tasks. The only language dependent components 
are the tokenizers and stop--lists. For both English and Spanish a 
stop--list is made up of all words that occur ten or more times 
in five randomly selected word training files of comparable size.  
All Duluth systems exclude the words in the stop--list from being 
features.

Each pair of systems is summarized below. All performance results
are based on accuracy  (correct/total) using fine-grained scoring. 
The name of the English system appears first, followed by the Spanish 
system.


{\bf Duluth1/Duluth6}
create an ensemble of three Naive Bayesian classifiers, 
where each is based on a different set of features. The hope is that 
these different views of the training examples will result in classifiers 
that make complementary errors, and that their combined performance will 
be better than any of the individual classifiers. 

Separate Naive Bayesian classifiers are learned from each
representation of the training examples. Each classifier assigns
probabilities to each of the possible senses of a test instance. 
These are summed and the sense with the largest value is used. This
technique is used in many of our ensembles and will be referred
to as a weighted vote. 

The first feature set is made up of bigrams, i.e., consecutive two word 
sequences, that can occur anywhere in the context with the ambiguous
word. To be selected as a feature, a bigram must occur two or more
times in the training examples and have a log-likelihood ratio 
($G^2$) value $\geq$ 6.635, which is associated with a p-value of .01. 

The second feature set is based on unigrams, i.e., one word sequences, 
that occur five or more times in the training data.  

The third feature set is made up of co-occurrence features that represent
words that occur on the immediate left or right of the target word. In 
effect, these are bigrams that include the target word. They must also 
occur two or more  times and have a log-likelihood ratio  $\geq$ 2.706, 
which is associated with a p-value of .10. 

These systems are inspired by \cite{Pedersen00b}, which presents an
ensemble of eighty-one Naive Bayesian classifiers based on varying sized 
windows of context to the left and right of the target word that define 
co-occurrence features. However, the current systems only use a three 
member ensemble to capture the spirit of simplicity and portability that 
underlies the Duluth approach to \textsc{Senseval-2}.

English accuracy was 53\%, Spanish was 58\%. 

{\bf Duluth2/Duluth7}
learn an ensemble of decision trees via bagging. 
Ten samples are drawn, with replacement, from the training
examples for a word. A decision tree is learned from each of these 
permutations of the training examples, and each of these trees 
becomes a member of the ensemble.
A test instance is assigned a sense based on a weighted vote among 
the members of the ensemble. 
In general decision tree learning can be overly influenced by a
small percentage of the training examples, so the goal of bagging
is to smooth out this instability.  

There is only one kind of feature used in these systems, bigrams that 
occur two or more times and  have a log-likelihood ratio $\geq$ 6.635. 
This is one of the three feature sets used in the Duluth1/Duluth6 
systems. 

The set of bigrams that meet these criteria become candidate features for  
the J48 decision tree learning algorithm, which is the Weka implementation
of the C4.5 algorithm. The decision tree learner first constructs a tree
of features that characterizes the training data exactly, and then
prunes features away to avoid over--fitting and allow it to generalize 
to the previously unseen test instances. Thus, a decision tree learner
performs a second cycle of feature selection and is not likely to use
all of the features that we identify prior to learning with BSP. The 
default C4.5 parameter settings are used for pruning. 

These systems are an extension of \cite{Pedersen01b}, which learns a
single decision tree where the representation of context is based on 
bigrams. This earlier work does not use bagging, and the top 100 bigrams 
according to the log-likelihood ratio are the candidate features. 

English accuracy was 54\%, Spanish was 60\%. 

{\bf Duluth3/Duluth8}
rely on the same features as 
Duluth1/Duluth6, but learn an ensemble of three bagged 
decision trees instead of  an ensemble of Naive Bayesian 
classifiers. There is a strong contrast between these techniques, 
since decision tree learners attempt to characterize the training
examples and find relationships among the features, while a Naive
Bayesian classifier is based on an assumption of conditional
independence among the features. 

The feature set used in these systems is from Duluth1/Duluth6
and consists of bigrams, unigrams and co-occurrences. A bagged decision 
tree is learned for each of the three kinds of features. 
The test instances are classified  by each of the bagged decision trees, 
and a majority vote is taken among the members to assign senses to the 
test instances. 

These are the most accurate of the Duluth systems for both English 
(57\%) and Spanish (61\%). These are within 7\% of the most accurate 
overall approaches for English (64\%) and  Spanish (68\%). 


{\bf Duluth4/Duluth9}
uses a Naive Bayesian classifier based on a bag of
words representation of context, where each unigram that occurs in 
the training data is taken as a feature. 
This is a common benchmark in word sense disambiguation studies and text 
classification problems. 

In the English training examples any word that occurs five or more times 
is used as a feature, and in the Spanish data any word that occurs two or 
more times is used. These  features are used to estimate the parameters of 
a Naive Bayesian  classifier. This will assign the most probable sense to 
a test instance, given the surrounding context. 

Accuracy for English was 54\%, and for Spanish 56\%. This Naive
Bayesian classifier was one of the three member classifiers in the
ensemble approach of Duluth1/Duluth7, which was 1\% less accurate 
for English and and 2\% more accurate for Spanish. 

{\bf Duluth5/Duluth10}
add a co--occurrence feature to the Duluth2/Duluth7
systems. In every other respect they are identical. The co-occurrence
feature was also used in Duluth1/Duluth6, and 
is essentially a bigram where one of the words is the 
ambiguous word.  These must occur two or more times in the
training examples and have a log-likelihood ratio $\geq$ 
2.706 to be included as a feature. 
In addition to the co-occurrence feature the bigram
feature from  Duluth2/Duluth7 is used, where a bigram must occur two or 
more times and have a log-likelihood ratio $\geq$  6.635. 

Accuracy for English was 55\%, and for Spanish 61\%. This was a slight
improvement over Duluth2 (54\%) and Duluth7 (60\%). 

{\bf DuluthA/DuluthX} 
build an ensemble of three different classifiers that are 
induced from the same  representation of the training examples. A weighted 
vote is taken to assign senses to test instances. The three classifiers 
are a bagged J48 decision tree,  a Naive Bayesian classifier, and the 
nearest neighbor classifier IB$k$,  where the number of neighbors 
parameter $k$ is set to 1.  

The context in which the ambiguous word occurs is represented by bigrams 
that may include zero, one, or two intervening words that are ignored.
To be considered as features these bigrams must occur two or more times 
and have a log-likelihood ratio $\geq$ 10.827, i.e., a p-value of 
.001. The log-likelihood ratio threshold is set to 0 for the Spanish data 
due to the smaller volume of data. 

English accuracy was 52\%, Spanish was 58\%. 

{\bf DuluthB/DuluthY}
are identical to Duluth5/Duluth10, except that rather
than learning an entire decision tree they stop the learning process
once the root of the decision tree is selected. The resulting one
node decision tree is called a decision stump. At worst a decision
stump will reproduce the most common sense baseline, and may do 
better if the selected feature is particularly informative. In 
previous work we have observed that decision stumps can serve as 
a very aggressive lower bound on performance \cite{Pedersen01b}. 

Decision stumps are the least accurate method for both English (DuluthB,  
51\%) and Spanish (DuluthY, 52\%), but are more accurate than the most 
common sense baseline for English (48\%) and Spanish (47\%). 

{\bf DuluthC/DuluthZ}
take a kitchen sink approach to ensemble creation, 
and combine the seven systems for English and Spanish into 
ensembles that assign senses to test instances by taking a weighted
vote among the members. 


Accuracy for English was 55\%, and for Spanish 59\%. This is less
than the accuracy of some of the members systems, suggesting
that the members of the ensemble are making redundant errors. 
 
\section{Discussion}

There are several hypotheses that underly and motivate these systems.

\subsection{Features Matter Most}

This hypothesis is at the core of much of our recent work. It 
holds that variations in learning algorithms matter far less to 
disambiguation performance than do variations in the features used to 
represent the context in which an ambiguous word occurs. In other words, 
an informative feature set will result in accurate disambiguation when 
used with a wide  range of learning algorithms, but there is no learning 
algorithm that can  overcome the limitations of an uninformative or 
misleading set of features. 

There are a number of demonstrations that can be made from the Duluth
systems in support of this hypothesis, but perhaps the clearest is found 
in comparing the systems Duluth1/Duluth6 and Duluth3/Duluth8. The first 
pair learns three Naive Bayesian classifiers  and the second learns three 
bagged decision trees. Both use the same feature set to represent the 
context in which ambiguous words occur.  There is a 3\%  improvement in
accuracy when using the decision trees. We believe this modest improvement 
when moving from a simple learning algorithm to a more complex one 
supports the hypothesis that the true dividends are to be found in 
improving the feature set. 



\subsection{50/25/25 Rule}

We hypothesize a 50/25/25 rule for supervised approaches to word sense 
disambiguation. This loosely holds that given a classifier learned
from a sample of sense--tagged training examples, about half of the
test instances are easily disambiguated, a quarter are harder but still 
possible, and the remaining quarter are extremely difficult. 
This is a minor variant of the 80/20 rule of time management, which holds 
that 20\% of effort accounts for 80\% of results. 

When the two highest ranking systems in the official English lexical 
sample results are compared there are 2180 test instances (50\%) that both 
disambiguate correctly using fine-grained scoring. There are an 
additional 1183 instances (28\%) where one of the two systems are 
correct, and 965 instances (22\%) that neither system can resolve. If 
these two systems were optimally combined, their accuracy would be 
78\%. If the third-place system is also considered, there are 1939 
instances (44.8\%) that all three systems can disambiguate, and 816 
(19\%) that none could resolve. 

For all the Duluth systems for English, there are 1705 instances 
(39\%) that all eight systems got correct. There are 1299 instances
(30\%) that none can resolve. The accuracy of an optimally combined
system would be 70\%. The most accurate individual system is Duluth3 
with 57\% accuracy.

For the Spanish Duluth systems, there are 856 instances (38\%) that all 
eight systems got correct. There are 478 instances (21\%) that none of 
the systems got correct. This results in an optimally combined result 
of 79\%. The most accurate Duluth system was Duluth8, with 1369 
correct instances (62\%). If the top ranked Spanish system (68\%) and  
Duluth8 are compared, there are 1086 instances (49\%) where both
are correct, 737  instances (33\%) where one or the other is correct,
and 402 instances (18\%) where neither system is correct. 

This is intended as a rule of thumb, and suggests that a fairly  
substantial percentage of test instances can be resolved by almost 
any means, and that a hard core of test instances will be very
difficult for any method to resolve.





\section{Acknowledgments}

This work has been partially supported by a National Science Foundation 
Faculty Early CAREER Development award (\#0092784).

The Bigram Statistics Package and SenseTools have been implemented by  
Satanjeev Banerjee.

%
%


\end{document}